\documentclass[letterpaper, 10 pt, conference]{ieeeconf}  

\IEEEoverridecommandlockouts                              
\usepackage{mathtools}
\usepackage{bbding}
\usepackage{graphicx}
\usepackage{caption,subcaption}
\usepackage{color, colortbl}
\definecolor{rcol}{gray}{0.9}
\usepackage[svgnames]{xcolor}

\newcommand{\mcrot}[4]{\multicolumn{#1}{#2}{\rlap{\rotatebox{#3}{#4}~}}}

\title{\LARGE \bf
Decoder Fusion RNN:\\
Context and Interaction Aware Decoders for Trajectory Prediction
}

\author{Edoardo Mello Rella$^{1}$, Jan-Nico Zaech$^{1}$, Alexander Liniger$^{1}$, and Luc Van Gool$^{1,2}$
\\\small\texttt{\{edoardo.mello-rella, jan-nico.zaech, alex.liniger, vangool\}@vision.ee.ethz.ch}%
\thanks{$^{1}$Computer Vision Lab, ETH Z\"uurich, Switzerland}
\thanks{$^{2}$PSI, KU Leuven, Belgium}
}

\begin{document}

\maketitle
\thispagestyle{empty}
\pagestyle{empty}

\begin{abstract}

Forecasting the future behavior of all traffic agents in the vicinity is a key task to achieve safe and reliable autonomous driving systems.
It is a challenging problem as agents adjust their behavior depending on their intentions, the others' actions, and the road layout.
In this paper, we propose Decoder Fusion RNN (DF-RNN), a recurrent, attention-based approach for motion forecasting. Our network is composed of a recurrent behavior encoder, an inter-agent multi-headed attention module, and a context-aware decoder. We design a map encoder that embeds polyline segments, combines them to create a graph structure, and merges their relevant parts with the agents' embeddings. We fuse the encoded map information with further inter-agent interactions only inside the decoder and propose to use \emph{explicit training} as a method to effectively utilize the information available. We demonstrate the efficacy of our method by testing it on the Argoverse motion forecasting dataset and show its state-of-the-art performance on the public benchmark.
\end{abstract}

\section{INTRODUCTION}

With autonomous driving obtaining increasing importance, it is paramount to develop systems able to drive accurately and safely. One requirement for an autonomous vehicle is to understand the context and the dynamic behavior of all the traffic actors while planning their paths.
Motion forecasting is a fundamental building block to achieve this: it consists of predicting the future trajectory of each surrounding traffic agent after observing a segment of its past motion. One possible approach is designing a neural network able to model and learn the principles that need to be followed in traffic.
This requires an architecture that can model how agents influence each other and how they interact, but that can also effectively use the information about the road structure and the traffic rules that apply to it. Finally, the network should also account for multiple possible behaviors to improve safety and accuracy.

To tackle this task, we propose an end-to-end trainable decoder-centered network designed to address all the aforementioned challenges. We use a recurrent architecture that auto-regressively predicts positions and velocities. On top of this structure, we exploit the attention mechanism to both model interaction between agents and to include road structure information. Given that driving is a dynamic task that requires reaction and adaptation to the others' choices, we find it important to allow contextual information to modify each agents' behavior at multiple points during the trajectory roll-out. Thus, we propose DF-RNN, a method that reduces the complexity of the early stages of the network by including the relevant information inside the recurrent decoder.

\begin{figure*}[th!]
  \centering
  \includegraphics[scale=0.6]{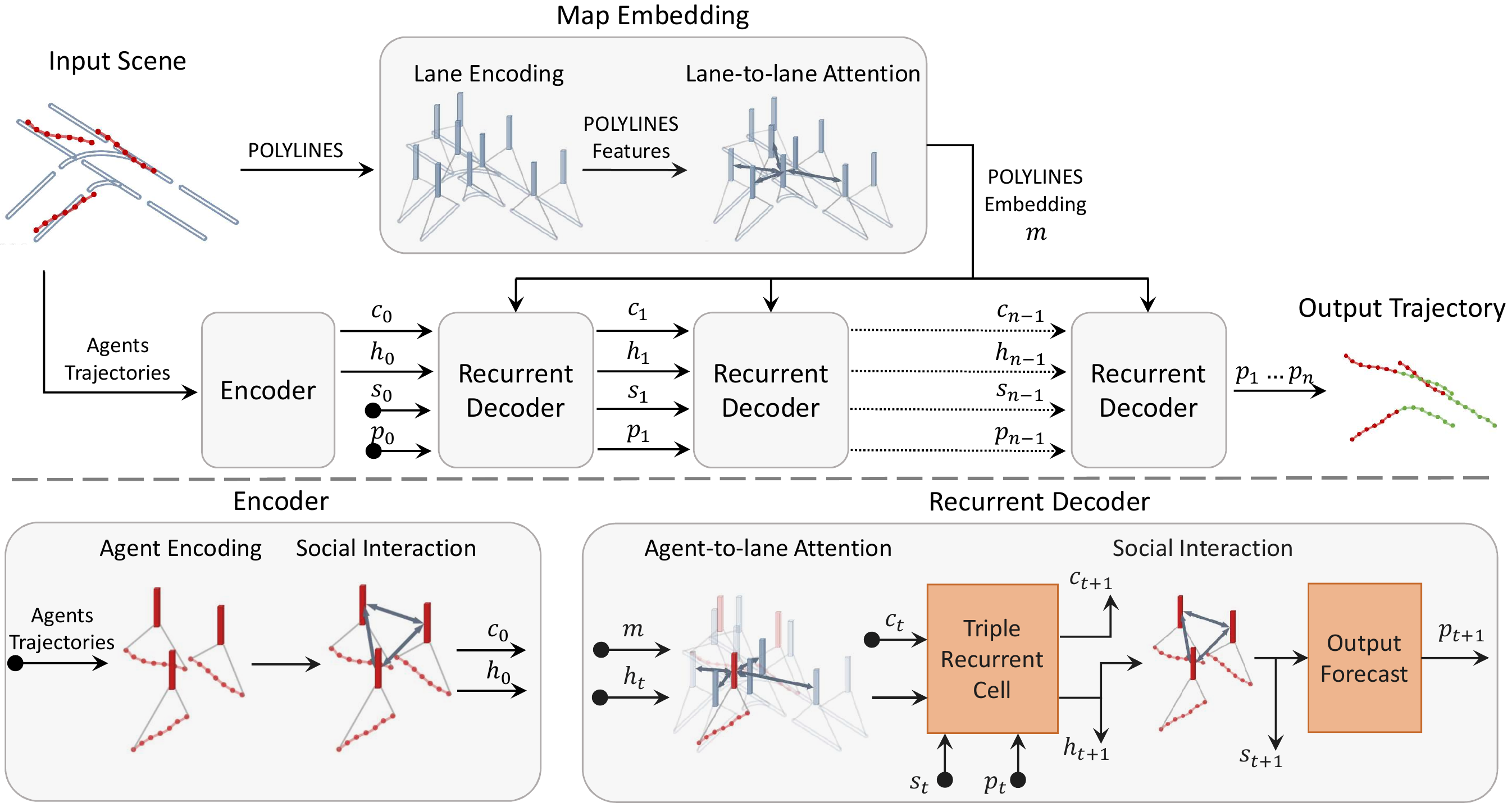}
  \caption{Overview of our network structure. An encoder produces embedding vectors for the agents and correlates them through social attention. Polylines are embedded and they are fused with the agents inside the decoder. The resulting vector is processed by the Triple Recurrent Cell and the social interaction layer before the output is finally predicted.}
  \label{fig:pipeline}
\end{figure*}

To successfully apply the outlined decoder-focused concept, we design several specific modules that are suited to add social and map context to the decoder. We also propose a novel recurrent cell, developed on top of an LSTM, that fuses position and social information in the decoder, called Triple Recurrent Cell (TRC). To model uncertainty and multi-modality, we use Gaussian mixture models and a soft winner-take-all (WTA) learning technique that favors diverse trajectory forecasting. Finally, we propose \emph{explicit training}, a technique that trains the complete network and two mini-networks, focusing respectively on social and map integration. By sharing weights between the three networks and training all at the same time, the complete network has to focus on the social and map interactions, but also avoid local minima where not all the information is fused.

Traditionally, motion forecasting has been solved by relying on rules and constraints based on human knowledge of the driving tasks \cite{markovprediction} \cite{pedestrianmodel} \cite{pedestrian2}; however, these methods suffer from being overly simple and not scalable.
Recent work has focused on learning-based neural network methods that have shown to outperform traditional ones. One possible approach has been to represent the trajectories as a Birds-Eye-View rasterized image with different channels for different time-steps \cite{rulesoftheroad2019} \cite{zaech2020action} \cite{covernet}. Representations like this can encode map information inside the image as well and process everything together with a convolutional neural network (CNN) architecture \cite{dsdnet} \cite{rulesoftheroad2019}.
Yet, this suffers from quantization errors and the high computational cost associated with processing rendered images.
 
A different line of work has used recurrent neural networks (RNN) to build a representation for the trajectories \cite{jean2019multihead} \cite{trajwithrecurrent}. On top of the recurrent architecture, contextual information is added to model the interaction between agents and to extract the road structure. \cite{sociallstm} uses the LSTM cell \cite{lstm} to encode agents and execute social pooling to correlate their behavior. \cite{socialgan} takes into account the multi-modality of predictions when modeling interactions between agents and exploits the power of generative adversarial networks (GAN) to improve the training. \cite{jean2019multihead} uses multi-headed attention \cite{attentionisallyouneed} as a model of interaction. \cite{sophiestanford} focuses on the interaction with other agents and the map using attention modules and GANs for training.

Recently, graph convolutional networks (GCN) have been used to unify interaction with other agents and with the environment into a single framework. \cite{vectornet} used a GCN to both embed high definition maps and trajectories as well as to model the relations between them. \cite{zhao2020tnt} builds on top of \cite{vectornet} but formulates predictions based on a set of target states reached by the trajectory to be forecasted. \cite{lanegcn} uses the connectivity information from the high definition maps to build a graph representation of the road and combines agents with the other traffic actors and the context. \cite{lanercnn} applies a GCN to encode agents' trajectories in relation to the centerlines and to model interactions and formulates predictions with a CNN. \cite{khandelwal2020whatif} uses a graph based attention approach and conditions the future prediction on the intended path of the agents.

Our work belongs to the class of RNN and GCN methods and uses an architecture that employs attention mechanisms to jointly model interactions between agents and with polylines generated from the map.
Furthermore, we apply lane-to-lane attention to add graph structure awareness to the lanes' embedding. Thanks to the use of social and map contextual information inside the recurrent decoder, it can generate an interactive influence on the agents, allowing to learn more complex interactions.

Several techniques have been proposed to generate multi-modal predictions. Some methods sample from a conditional variational auto-encoder (CVAE) \cite{rulesoftheroad2019} \cite{rhinehart2018r2p2} \cite{casas2020implicit} while others directly regress multi-modal parameters \cite{cui2019multimodal} \cite{jean2019multihead}. In our work, we use the second approach and regress the parameters of a mixture of Gaussians at each time-step.

To achieve realistic and diverse trajectories and to exploit the map information provided with the data, several training techniques have been developed. \cite{wta_initial} proposes to use the WTA approach and \cite{evolvingwta} changes it to an evolving WTA that progressively reduces the number of "winning" modes during training. \cite{sun2020reciprocal} assumes that trajectories should be forward and backward predictable and uses reciprocal attacks to enforce realistic predictions. In the field of imitation learning, \cite{chaffeurnet} drops out the past trajectory for the agents in 50\% of the training cases to learn to rely on the map information. Tackling the problem of multi-modal classification, \cite{multihard} shows the benefit of adding supervision to each input modality and optimizes the weights of each loss term. In our work, developing on top of these methods, we apply a soft version of WTA and adapt the concept behind \cite{multihard} to improve the level of context-awareness in our model.

In summary our contributions are threefold:
\begin{itemize}
    \item A novel decoder-focused recurrent motion forecasting architecture that fuses map and social interactions in the decoder stage, not only in the encoder.
    \item The Triple Recurrent Cell used in the decoder to merge the agent's hidden representation with the interaction contribution and the previously predicted position.
    \item \emph{Explicit training}, a training technique that uses multiple parameter-sharing networks, to force the combined network to rely on map and social information.
\end{itemize}

\section{METHOD}

In this section, we introduce our decoder-focused network architecture and provide details about the training technique that prompts it to use all available information sources.

\subsection{Network Structure}
\label{sec:NS}

To achieve accurate predictions in motion forecasting, it is essential to jointly exploit the information provided about the past trajectory of each agent and the surrounding environment. Therefore, it is necessary to design a network architecture able to embed each of these inputs and to combine them in an effective way. We assume that modeling short-term dependencies simplifies this learning task and that each agent should continuously refine its behavior based on the changing contextual information; it is easier to learn how agents interact with each other in the next second rather than two seconds into the future. Furthermore, including contextual information at multiple steps within the prediction stage can model complex relations and compensate for possible inaccuracies. Therefore, we move the bulk of the model to the decoder stage so that, at multiple prediction steps, agents interact with each other and with the environment.

Our model has an encoder-decoder structure, where the encoder generates a feature representation of each agent and the decoder recursively predicts trajectories. The encoder uses convolutions along time, an LSTM cell \cite{lstm} to combine all time-steps into a single vector and an interaction layer to model the influence that agents have on each other. Similarly, the decoder uses a recurrent structure to roll out predictions for all future time-steps. Furthermore, the architecture contains a map embedding module and an agent-to-lane as well as an agent-to-agent attention layer. Our complete network pipeline is shown in Figure \ref{fig:pipeline}, and the individual building blocks are explained next.

\textbf{Social Interaction:} Similar to previous work \cite{jean2019multihead}, we model the interaction between agents with multi-headed attention \cite{attentionisallyouneed}. Each agent is represented with an embedding vector that encodes its observed trajectory and is used to create a key $K$, query $Q$, and value vector $V$ for each of the $H$ attention heads. With each attention head that can recover interaction patterns, such as focusing on the traffic agent in front, in the back, or in some other area \cite{jean2019multihead}, we set $H = 8$ to cover a sufficient number of directions. Inside each attention head, the level of interaction between the agents in the scene is modeled with the dot product between the key and query vector. This set of agent-specific scores is then used to weight the combination of the value vectors
\begin{align}
    O = \text{softmax}\left(\frac{QK^T}{\sqrt{d_k}}\right)V, \label{eqn:attention}
\end{align}
with $d_k$ the key dimension. Each head outputs a vector for each agent that constitutes a representation of the social interaction influence on it. The results of all the attention heads for an agent are linearly fused and summed to the original embedding which is propagated with a skip connection.

\textbf{Lane-to-lane Attention:} 
The road structure is represented with polylines, which are ordered sequences of points on the street centerline. If combined, multiple polylines can cover the whole road structure in a scene and lane-to-lane attention can be used to combine them.
The input to the map attention module is obtained using a set of convolutions applied to the original polyline structure. Using two convolutional layers and a weighted averaging, the polylines are converted to 64-dimensional feature vectors.
These representations are then fed into a multi-headed attention module to obtain an embedding for each polyline that combines information about itself and its neighborhood.
In particular, the similarity scores between them - as computed inside inter-agent attention - can be seen as the elements of an adjacency matrix representing a graph that has polylines as nodes.

\textbf{Agent-to-lane Attention:}
After lane-to-lane attention, an agent-specific map representation is computed using single-headed agent-to-lane attention, which can focus primarily on the part of the road useful for prediction.
We first compute a key vector for each agent and query and value vectors for the lane embeddings. Then, the attention mechanism is applied to these vectors as shown in Equation \ref{eqn:attention}. In parallel, a skip connection preserves the agent's embedding which is added to the map contribution.
To stabilize the training process, the influence of polylines is discounted with a direction-aware distance between the agent and the polyline.
This is based on the assumption that regions of the road close to the agent and with the same direction should have a higher influence on its behavior than other areas.

\textbf{Interactive Decoder:} 
We can now introduce our novel interactive fusion decoder, which uses both map and interaction information. This allows the network to always use up-to-date information, reducing the dependency on the encoder and taking correct short term decisions inside the decoder.

The inputs to the decoder for each agent are its embedding vector, its current position, and its social interaction influence. The embedding vector is a feature representation for the agent, obtained by encoding the observed trajectory and applying the first interaction layer to it. The current position is either the last observation or the previous prediction, and the interaction influence is the output of an inter-agent social attention module executed at each decoder step.

First, the embedding vector is combined with the road layout using the agent-to-lane attention architecture. To minimize the computational overhead of adding the map, this is executed only once every $n$ prediction steps, with $n$ being a dataset-dependent value. More specifically, considering an urban environment, we set $n$ so that map information is integrated once every second of forecasting. The map-aware embedding, the current position, and the interaction influence are the inputs to the TRC, which generates the new agent's embedding.
The result is then processed with a social interaction module that outputs a vector, which is used as the social interaction signal for the TRC in the next time-step, and as input to three fully connected layers that compute the new prediction.

\begin{figure}[t!]
  \centering
  \includegraphics[scale=0.35]{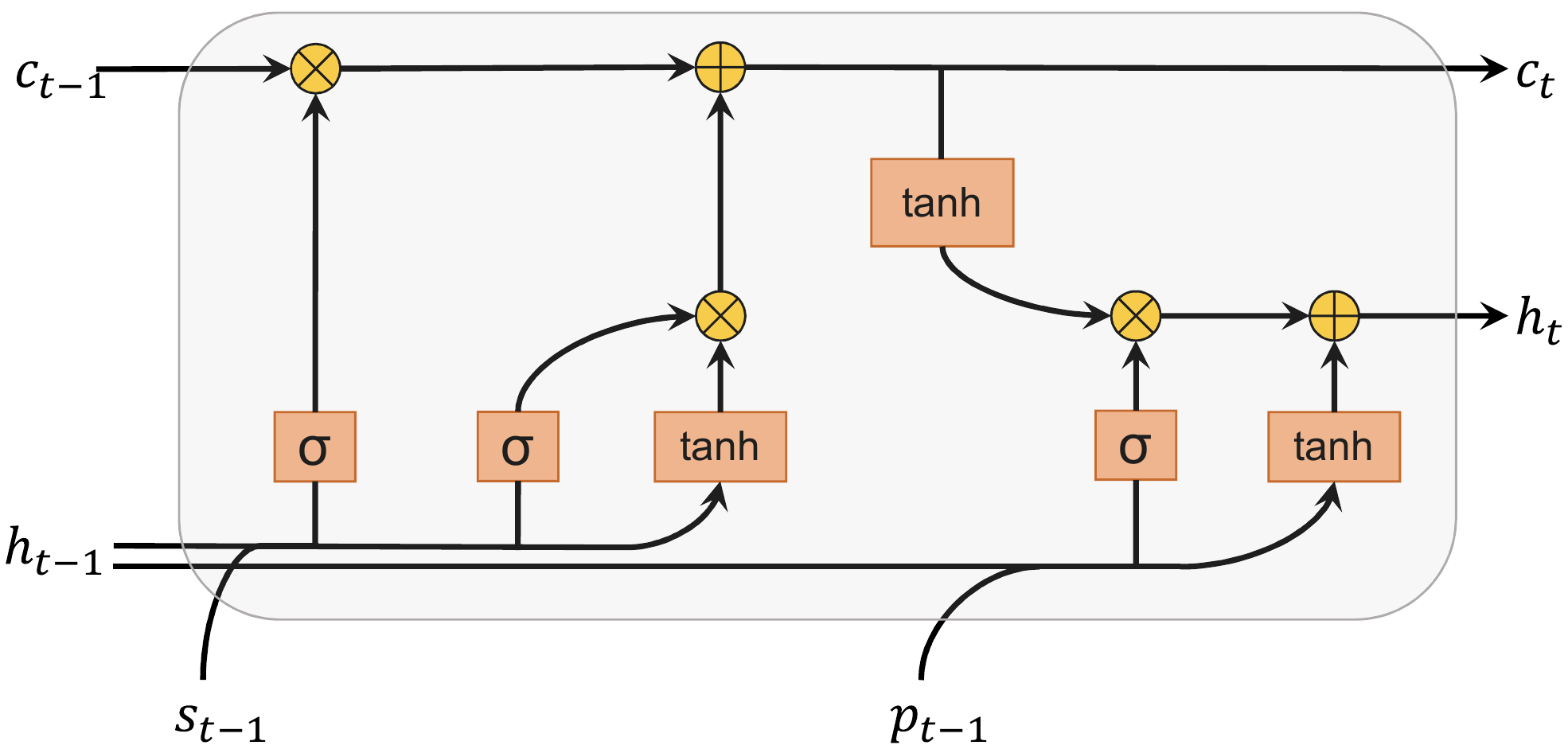}
  \caption{Triple Recurrent Cell: it combines the agent's embedding $h$, the social interaction signal $s$ and the current position $p$ with the cell state $c$.}
  \label{fig:rc}
\end{figure}

\textbf{Triple Recurrent Cell:} The recurrent cell inside the decoder has three inputs and a state signal - initialized as the agent's embedding from the encoder stage - and, therefore, no default module can be used. We design the novel TRC for this purpose that adapts the LSTM cell to have an additional input. As shown in Figure \ref{fig:rc}, the TRC implements the following equations, using the notation below:
\begin{itemize}
    \item We call $h$ the agent's embedding, $c$ the cell state, $s$ the social interaction input, and $p$ the position input.
    \item Capital letters are adopted for matrices and $k_x$ for an additive constant in part $x$.
    \item The symbol $\circ$ indicates elementwise multiplications.
    \item The current time-step is denoted with the subscript $t$.
\end{itemize}
\begin{align}
    i_t &= \sigma(H_1 h_{t-1} + S_1 s_t + k_a)
\label{eqn:cell1}\\
    \tilde{c}_t &= \tanh(H_2 h_{t-1} + S_2 s_t + k_b)
\label{eqn:cell2}\\
    f_t &= \sigma(H_3 h_{t-1} + S_3 s_t + k_d)
\label{eqn:cell3}\\
    c_t &= c_{t-1} \circ f_t + i_t \circ \tilde{c}_t
\label{eqn:cell4}\\
    o_t &= \sigma(H_4 h_{t-1} + P_1 p_t + k_i)
\label{eqn:cell5}\\
    u_t &= \tanh(H_5 h_{t-1} + P_2 p_t + k_j)
\label{eqn:cell6}\\
    h_t &= \tanh(c_t) \circ o_t + u_t
\label{eqn:cell7}
\end{align}
As in an LSTM cell, the cell state $c_t$ is updated based on the agent embedding $h_t$ and an additional signal - the social interaction $s_t$ - using a forget gate (Eqn. \ref{eqn:cell3}) and an input gate (Eqn. \ref{eqn:cell1}). The result of Equation \ref{eqn:cell4} is then the new cell state $c_t$ that is used in the next decoder step. It is also used to compute the cell output $h_t$, combining it with the agent's embedding and the current position, through the output gate (Eqn. \ref{eqn:cell5}) and an additional updating gate (Eqn. \ref{eqn:cell6}).

The idea behind the TRC is that the cell state should encode the high level behavior of each agent, while the output needs to know the current position to update it. Thus, the cell state is updated only using the agent's embedding and the social interaction information, while the current position modifies the output used to regress the new prediction.

\subsection{Input and Output}
\label{sec:IaO}

After specifying the architecture used for the task, it is necessary to better define the form of the input and output.
The data used consists of a sequence of observed positions for each agent and the model has to predict output sequences accordingly.
The positions are defined as $(X, Y)$ coordinates in a scene coordinate system centered at the last observation of a specific anchor agent - the choice of the anchor agent is dataset specific and will be discussed later.
The original input sequence is smoothed with a bidirectional Kalman filter used also to augment the agent's information with estimated velocity and acceleration and to interpolate the position sequences when steps are missing; these gaps in the sequence can occur due to occlusions during the dataset recording. Furthermore, to reduce the negative effect of unevenly sampled data, the time-step value is used as an additional feature in the input.

\textbf{Velocity Integrator:} Given the input, the network needs to output the future trajectory. To do it, the network regresses velocities in the $X$ and $Y$ direction that are fed to an integrator to retrieve the absolute positions in scene coordinates:
\begin{equation}
    p_t = p_{t-1} + T \cdot v_t\,.
\label{eqn:int1}
\end{equation}
The model assumes constant velocity during a time-step $T$ and $p$ is the position that is updated, while $v$ are the velocities. To simplify the backpropagation during training, we apply a loss both on the predicted positions and velocities.

\textbf{Multi-modal predictions:}
Motion forecasting is generally accepted as a task with intrinsic uncertainty and it is not sufficient to predict a single output trajectory.
To predict multiple trajectories with their relative probability, we model the different behaviors using Gaussian mixture models (GMM) as defined in Equation \eqref{eqn:gmix3}. The GMM is defined as the sum of multiple Gaussian distributions, weighted by the non-negative mixing parameters $w_i$ that sum to one $\sum_{i=0}^{I}{w_i} = 1$. In our work, we are interested in two types of distributions, the velocities $v = [v_X, v_Y]^T$ predicted by our network, and the positions obtained by integrating these velocities $p = [X,Y]^T$. Thus, we restrict our attention to Gaussian distributions with an $X$ and $Y$ component assuming that they are independent, which yields a diagonal covariance matrix $\Sigma$. The resulting GMM with mean $\mu$ and covariance $\Sigma$ is given in Equations \ref{eqn:gmix3} and \ref{eqn:gmix1},
\begin{align}
    &P(j) = \sum_{i=0}^{I}{w_i\mathcal{N}(\mu_i, \Sigma_i)}\label{eqn:gmix3}\\
    &\mu = \begin{pmatrix}
          \mu_x\\ 
          \mu_y
        \end{pmatrix}, \quad
    \Sigma = \begin{pmatrix}
          \sigma_x^2 & 0\\ 
          0 & \sigma_y^2
        \end{pmatrix}.\label{eqn:gmix1}
\end{align}

Note that given this definition we have two GMM; the position distribution $P_p$ which uses the components $\mu_p$, $\Sigma_p$, and $w_p$, and the velocity distribution $P_v$ that consists of $\mu_v$, $\Sigma_v$, and $w_v$.
Since both of the distributions are crucial for predicting the next time-step, all the parameters of the two are feedbacked as the position input to the TRC.
During evaluation, using probability distributions for the output is not suitable as it is not the standard representation. Therefore, each of the modes in the GMM is sampled to produce a sequence of discrete positions that form our trajectories.

\textbf{Loss Function:} To train the network, it is necessary to take into account the multi-modality of the prediction. For this reason, we use the negative log-likelihood loss between the predicted position and velocity distributions and the actual position and velocity.
To define our negative log-likelihood (NLL) cost, we first specify the target $z$, which consists both of the position and the velocities and is given as $z = [X,Y,v_X,v_Y]^T$. To make the notation simpler we combine our two GMM to one, by concatenating them in the same order as in the target $z$. This results in a new GMM with mean $\mu_z$, covariance $\Sigma_z$ and weights $w_z$. Thus, we can define the NLL between the ground truth (GT) target $\hat z$ and each mode of the GMM as:
\begin{align}
\begin{split}
    NLL(\hat z, \mu_z, \Sigma_z) = &\frac{1}{2}\left((\hat z - \mu_z)^T \Sigma_z^{-1} (\hat z - \mu_z)\right) \\
    &+ \ln \left(2\pi \sqrt{\text{det}\Sigma_z}\right).
\end{split}
\end{align}
Given the NLL for each mode of the GMM, we compute the loss by combining all the $I$ modes in the following way:
\begin{align}
\begin{split}
    \mathcal{L}&(\hat z, N_{z,i=1}^I) =\\
    & - \ln \left(\sum_{i=1}^I w_i \exp{(- NLL(\hat z, \mu_{z,i}, \Sigma_{z,i}))} \right).
\end{split}
\label{eqn:loss}
\end{align}

\subsection{Training Techniques}
\label{sec:TT}

A significant challenge in motion forecasting is properly training the network since it has to be invariant to the position and rotation of the agents in the scene, produce diverse modalities, and combine the agents' trajectories with the map information and inter-agent interactions.

To achieve the aforementioned invariance, we apply a random transformation, such as a combination of rotations and translations, to the training set. Invariance is also favored by predicting the future trajectory of all the agents in a scene as their relative position varies in the data.

\textbf{Soft Winner Takes All:} Predicting diverse modalities is a challenging task as, at each training step, all the forecasted modes are trained to produce the one correct trajectory, while the goal is that each one specializes to predict a different behavior. This can be partially favored by the Gaussian mixing parameter $w_i$ as the loss has a different weight depending on it. We propose to amplify this effect by applying a soft WTA approach that explicitly changes the weight of the loss based on the mode accuracy. The modalities that achieve the best result on a scene are strongly supervised to output the correct prediction while the others are less influenced.
This is achieved by multiplying the mixing component $w_i$ by the weight $a_i$ in Equation \ref{eqn:loss}.
The weight $a_i$ is computed based on the final displacement error $\mathcal{D}(p_i, \hat{p}_i)$, which is the Euclidean distance between the last position $p_{T,i}$, predicted sampling the $i$-th mode, and the GT final position $\hat p_T$. The weighting factor is then computed as:
\begin{align}
    & a_i = e^{m(1-\alpha)(\hat{d}-\mathcal{D}(p_i, \hat{p}_i))}, \label{eqn:wta}
\end{align}
where $m$ and $\hat{d}$ are weights that in our case are set to $m = 20$ and $\hat{d} = 2$. Finally, $\alpha$ regulates the strength of the difference and is annealed from $1$ to $0$ through the training epochs. When $\alpha \sim 1$ all the weights $a_i$ are equal, while as it approaches $0$ they vary strongly depending on the accuracy.

\textbf{Explicit Training:} Properly exploiting the information provided by the social interaction and road layout is another challenge that a strong training technique can help overcome. It is a hard problem as the two sources of information are partially overlapping and redundant. In fact, it is possible to infer part of the road structure from the position and trajectory of the agents in a scene. Therefore, the network can find the sub-optimal solution of removing map information.
To avoid this behaviour, we propose to augment the training using two sub-networks in addition to the complete one; we call this \emph{explicit training}.
The two sub-networks share the weights with the full one. The first sub-network only uses the modules exploiting social interaction and skips the agent-to-lane attention layer. The second sub-network only uses map information. More specifically, each interaction layer and the encoder are removed and the initial agent embedding $h_0$ is obtained applying a fully connected layer to the last observed position.
The complete network benefits from the \emph{explicit training} as its interaction and map addition modules undergo two levels of training.
Furthermore, by training the three networks simultaneously, the two small networks boost the use of the map and social information, and the combined network learns to fuse the information streams.

\section{EXPERIMENTS}

\begin{table}[thpb]
 \begin{center}
 \begin{tabular}{c | c c c c}
  \textbf{Method} & \textbf{minFDE $\downarrow$} & \textbf{minADE $\downarrow$} & \textbf{MR $\downarrow$} & \textbf{DAC $\uparrow$}\\ \hline
  Argoverse NN \cite{argoverse} & 3.287 & 1.713 & 0.537 & 0.868\\
  UULM-MRM & 1.55 & 0.96 & 0.22 & 0.98 \\
  SAMMP \cite{jean2019multihead} & 1.55 & 0.95 & 0.19 & 0.98 \\
  TNT \cite{zhao2020tnt} & 1.538 & 0.936 & 0.133 & 0.988\\
  LaneRCNN \cite{lanercnn} & 1.453 & 0.904 & 0.123 & 0.990\\ 
  WIMP \cite{khandelwal2020whatif} & 1.422 & 0.900 & 0.167 & 0.982\\
  Jean & 1.421 & 0.997 & 0.131 & 0.987\\ 
  LaneGCN \cite{lanegcn} & 1.364 & 0.868 & 0.163 & 0.980\\ \hline
  \textbf{Ours} & 1.414 & 0.958 & 0.159 & 0.986\\
 \end{tabular}
 \end{center}
 \caption{Results of state-of-the-art methods on Argoverse \cite{argoverse} test set. All results are reported on multimodal predictions with 6 different possible trajectories and are ordered according to \textit{minFDE}. \textit{Jean} and \textit{SAMMP} are the same method that was improved over time. The result of \textit{SAMMP} was reported as a winning method in Argoverse Forecasting Challenge 2019 while \textit{Jean} was the winning method of Argoverse Forecasting Challenge 2020.}
 \label{tab:SOTA}
\end{table}

\begin{figure*}[ht]
  \centering
  \begin{subfigure}[b]{0.25\linewidth}
    \framebox{\centering\includegraphics[width=103pt]{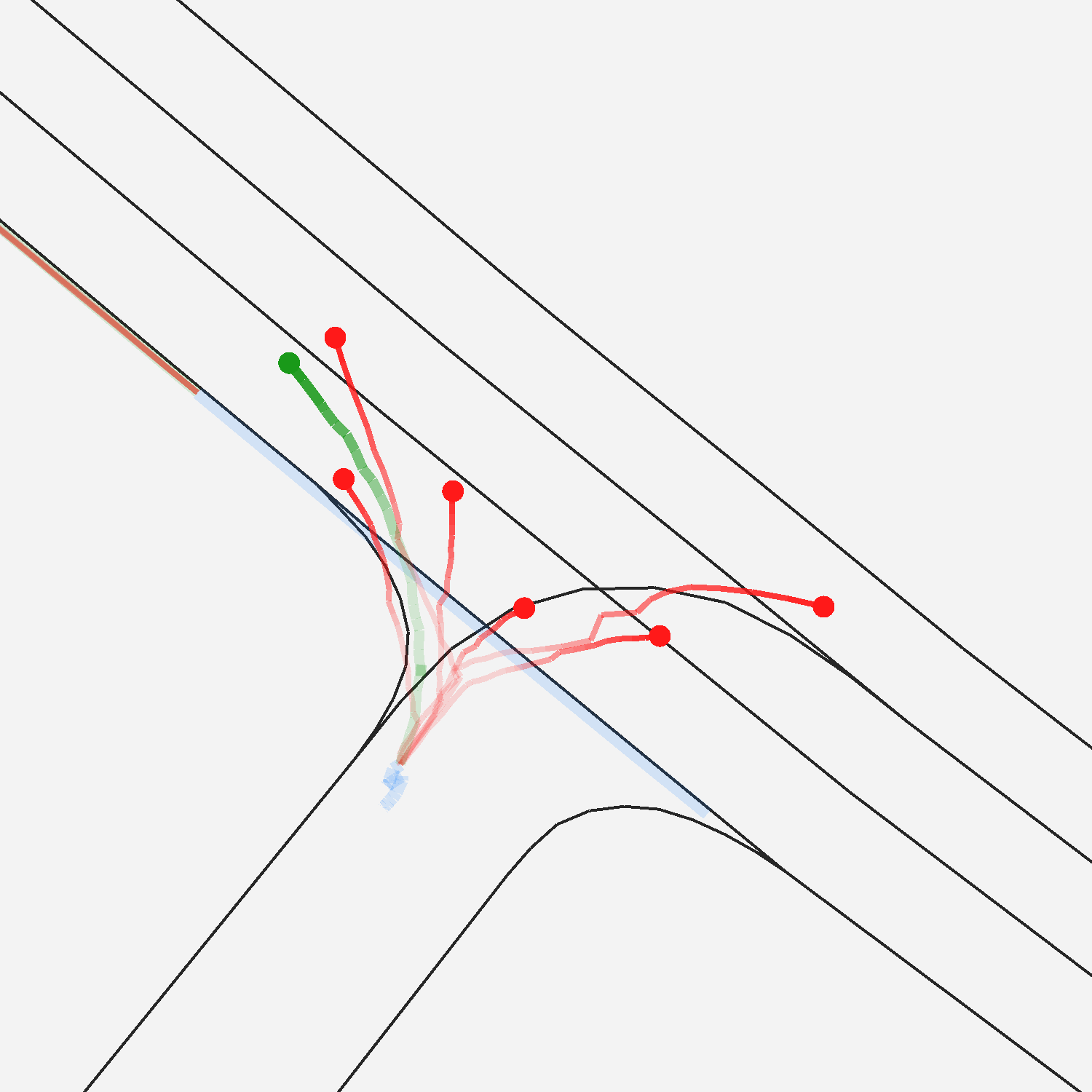}}
    \caption{\label{fig:s5}}
  \end{subfigure}%
  \begin{subfigure}[b]{0.25\linewidth}
    \framebox{\centering\includegraphics[width=103pt]{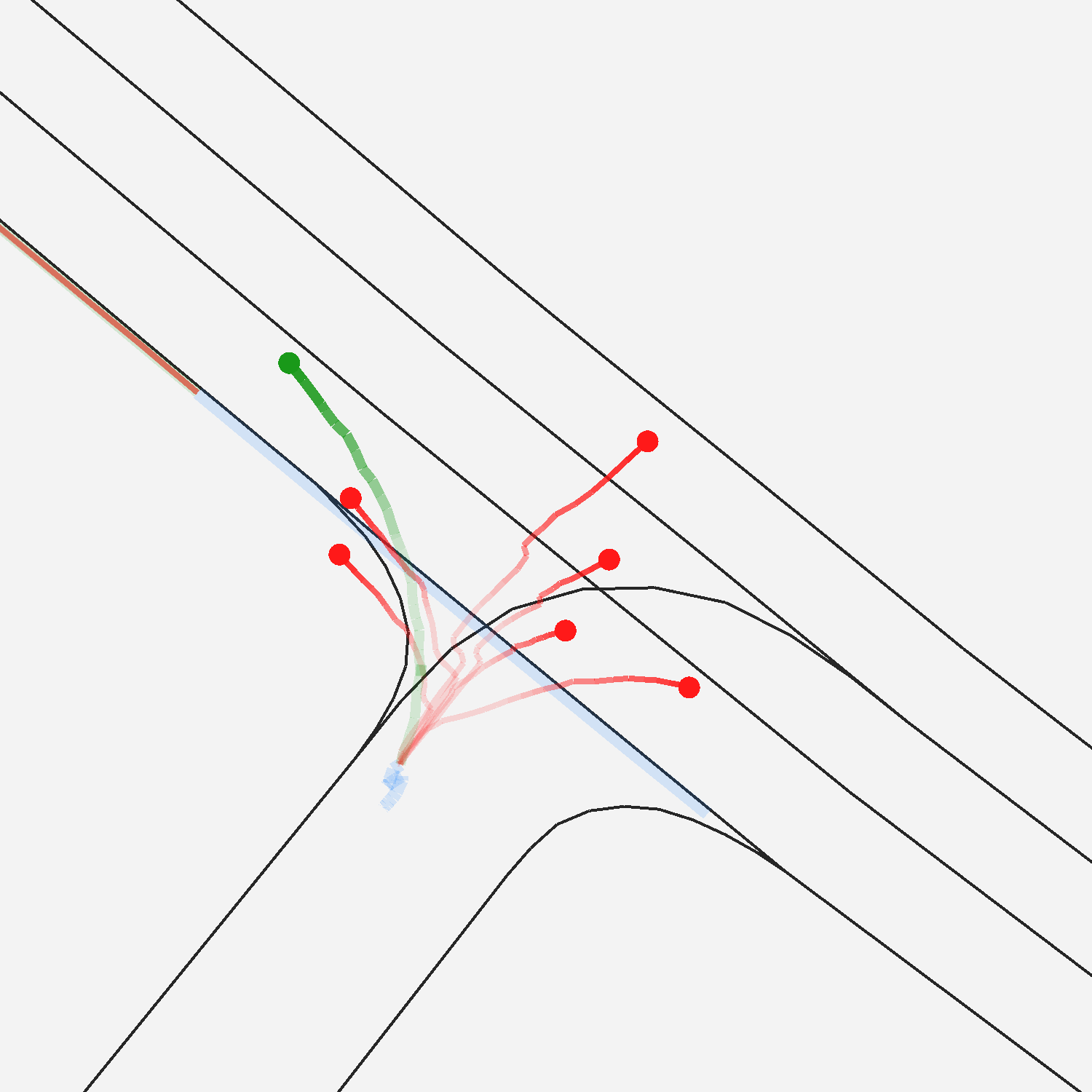}}
    \caption{\label{fig:s6}}
  \end{subfigure}%
  \begin{subfigure}[b]{0.25\linewidth}
    \framebox{\centering\includegraphics[width=103pt]{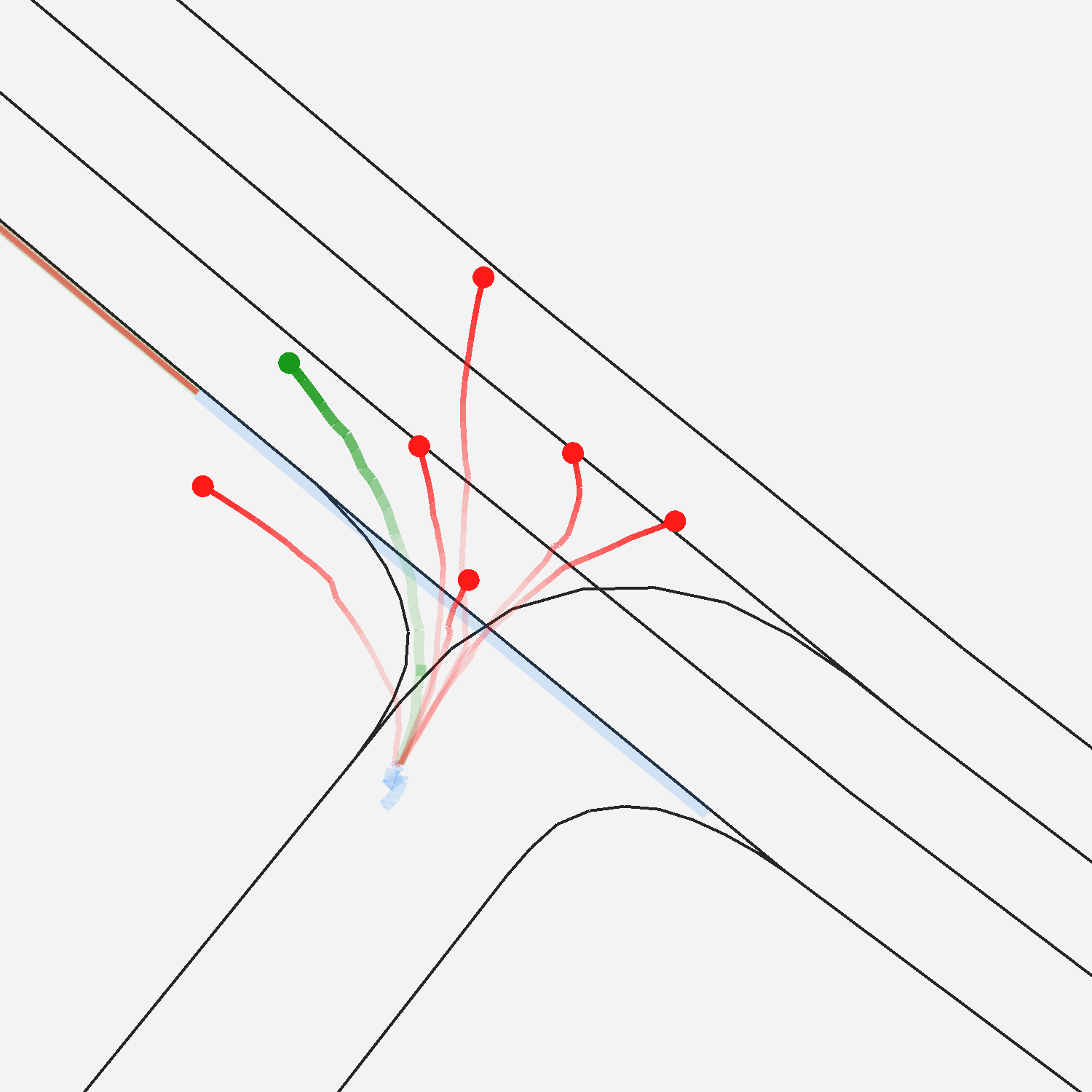}}
    \caption{\label{fig:s7}}
  \end{subfigure}%
  \begin{subfigure}[b]{0.25\linewidth}
    \framebox{\centering\includegraphics[width=103pt]{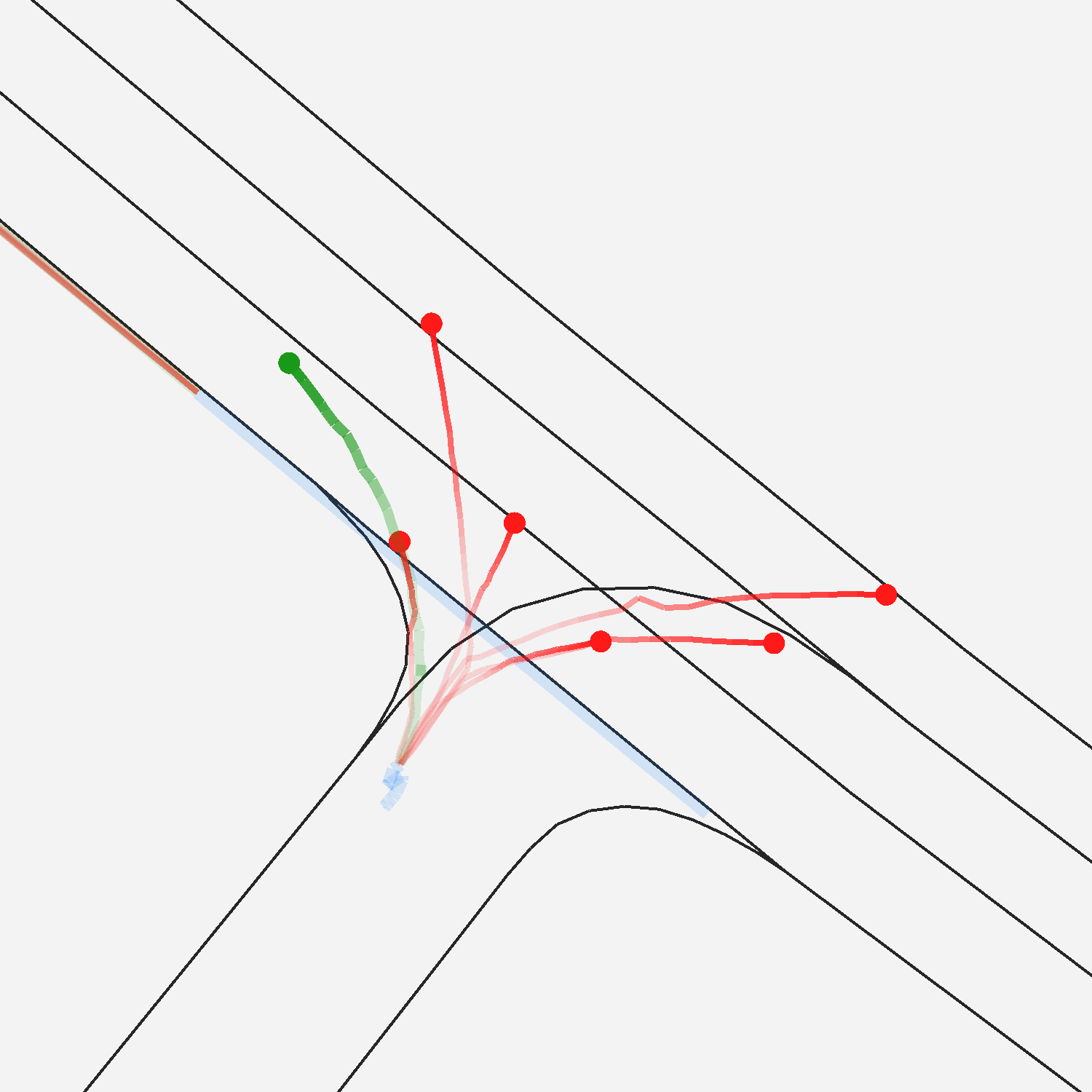}}
    \caption{\label{fig:s8}}
  \end{subfigure}
  \begin{subfigure}[b]{0.25\linewidth}
    \framebox{\centering\includegraphics[width=103pt]{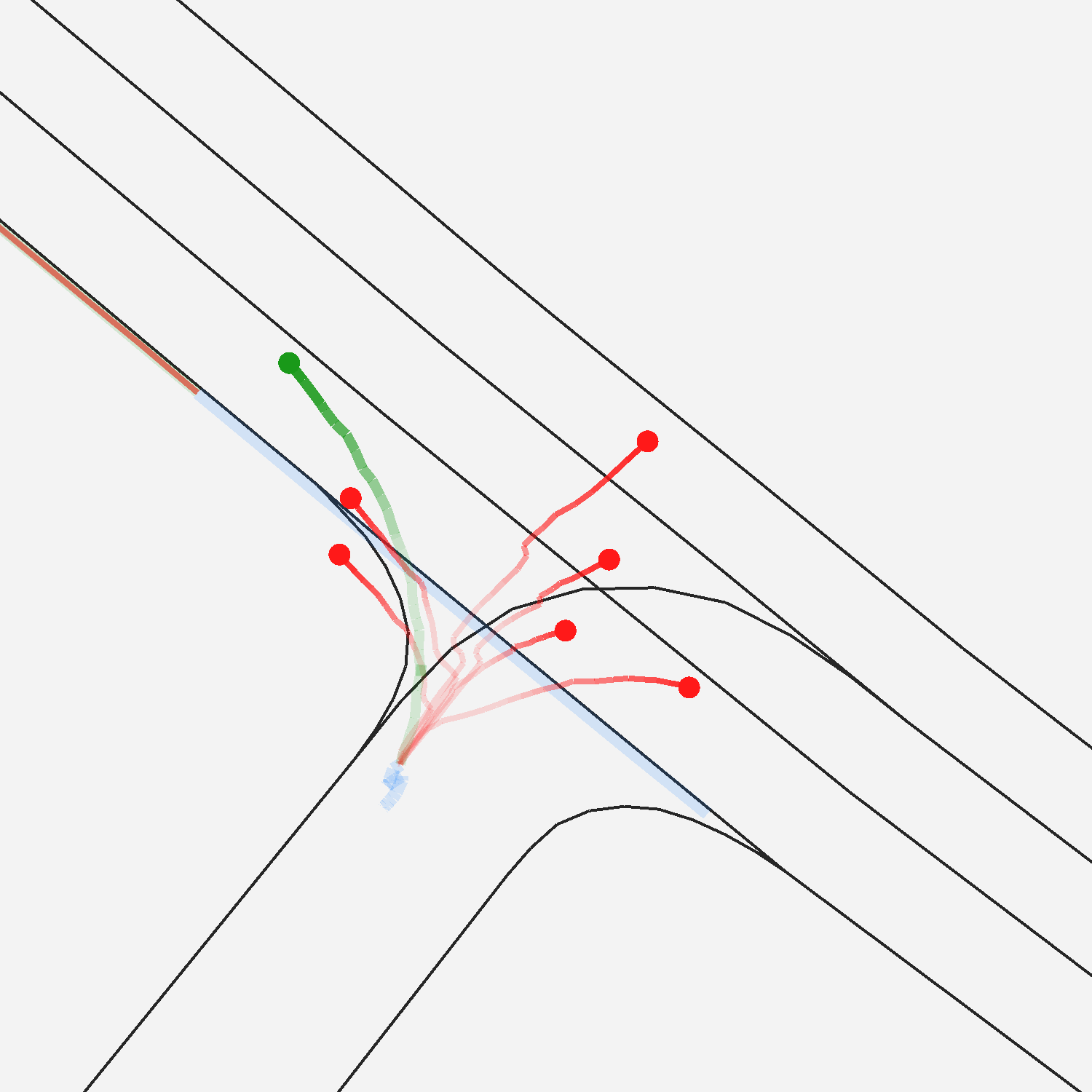}}
    \caption{\label{fig:s1}}
  \end{subfigure}%
  \begin{subfigure}[b]{0.25\linewidth}
    \framebox{\centering\includegraphics[width=103pt]{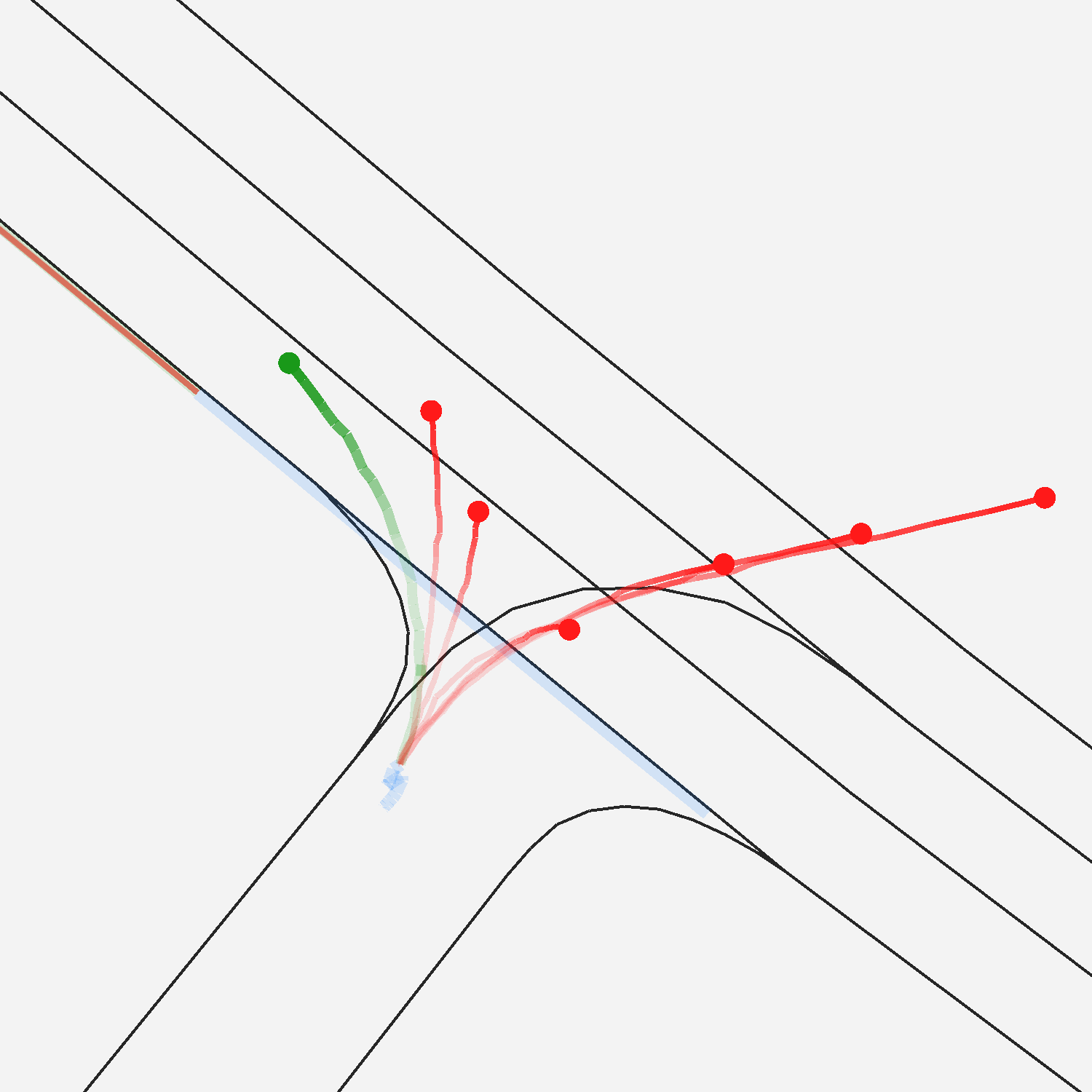}}
    \caption{\label{fig:s2}}
  \end{subfigure}%
  \begin{subfigure}[b]{0.25\linewidth}
    \framebox{\centering\includegraphics[width=103pt]{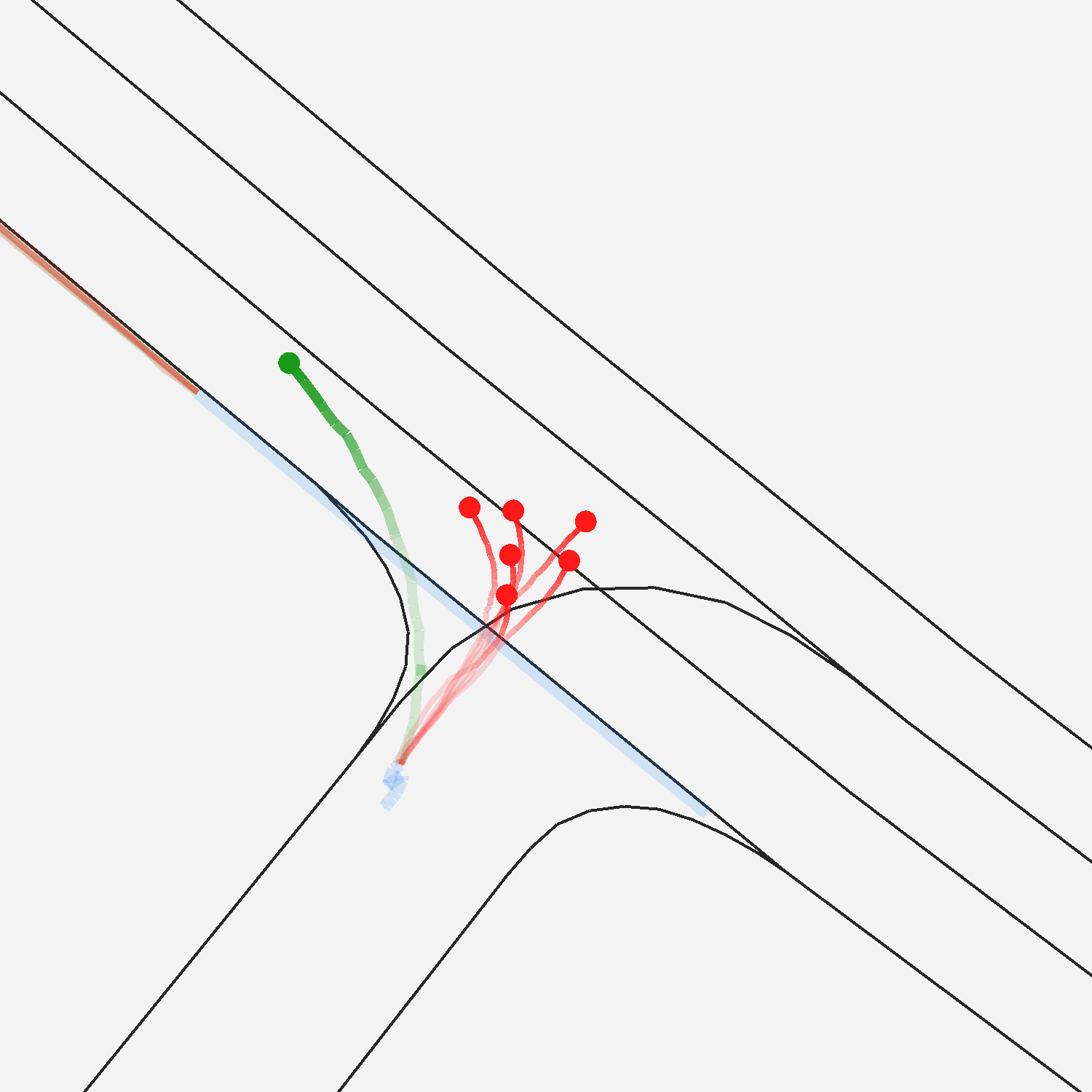}}
    \caption{\label{fig:s3}}
  \end{subfigure}%
  \begin{subfigure}[b]{0.25\linewidth}
    \framebox{\centering\includegraphics[width=103pt]{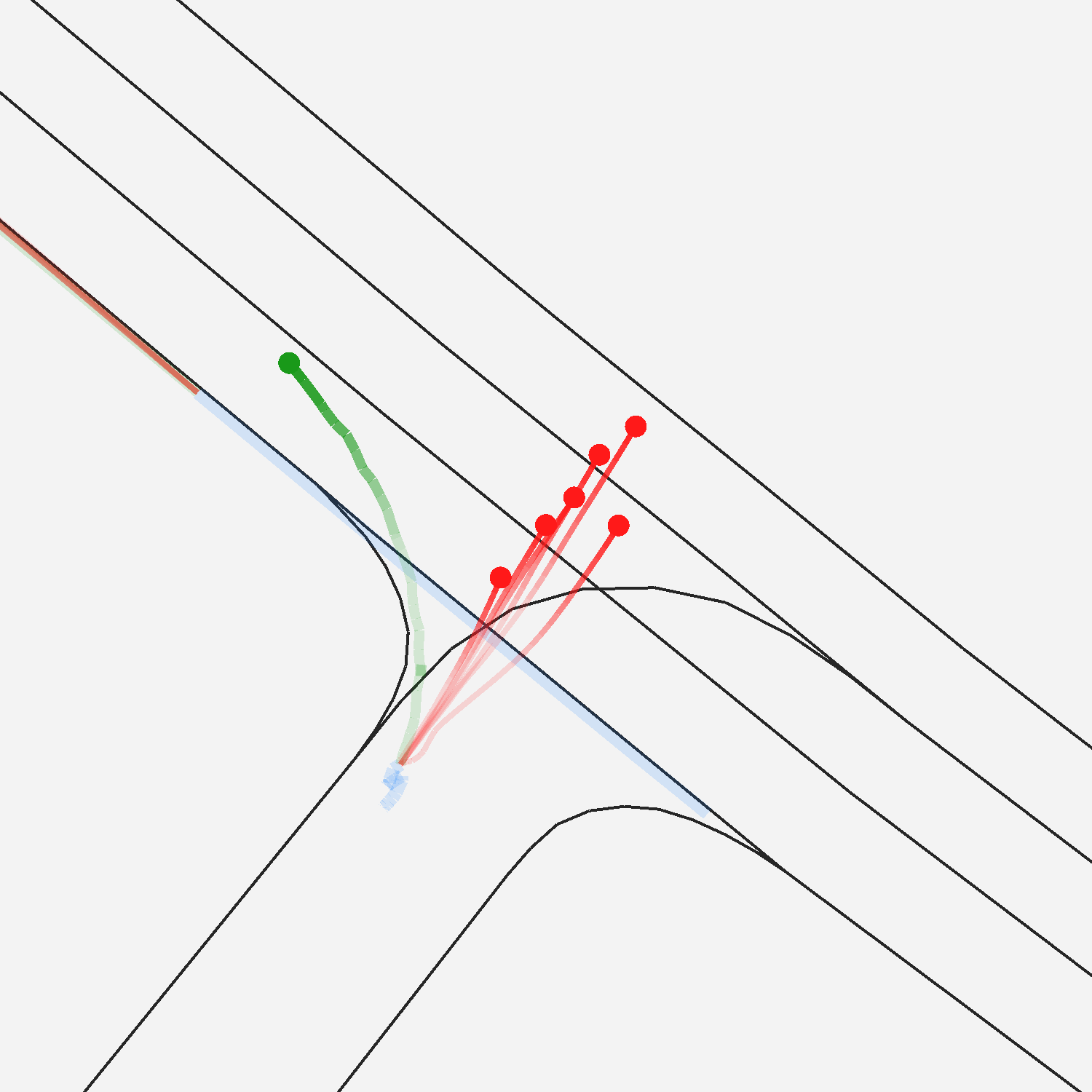}}
    \caption{\label{fig:s4}}
  \end{subfigure}
  \caption{Sample prediction on a hard example for the ablation study. The blue segment is the observed path, the green one is the GT future trajectory and the red ones are the network predictions. Figures \ref{fig:s5}, \ref{fig:s6}, \ref{fig:s7}, \ref{fig:s8} show the effect of progressively removing the use of \emph{explicit training}, random transformations and soft WTA from training. Figures \ref{fig:s1}, \ref{fig:s2}, \ref{fig:s3}, \ref{fig:s4} show the results of removing different modules from the complete network structure. In the order, the map information is moved from the decoder to the encoder, then it is completely removed and finally, also social interaction is ablated.}
  \label{fig:ablation}
\end{figure*}

In this section we present the experimental results of applying our method on a widely used motion forecasting dataset, Argoverse \cite{argoverse}. In Section \ref{section:DaM}, we give an overview of the dataset and the metric used to evaluate the models. In Section \ref{section:NA}, we evaluate the network in comparison to several recently proposed state-of-the-art methods. Lastly, in Section \ref{section:TM}, we execute an ablation study to evaluate the effect of our modules and the training techniques used.

\subsection{Dataset and Metric}
\label{section:DaM}

The Argoverse Motion Forecasting dataset \cite{argoverse} includes more than 300,000 5 second scenes, consisting of several agents tracked throughout the scene. The position of each traffic agent is identified with the center of mass of the objects without any distinction between car types, trucks, cyclists, and pedestrians. With 50 time-steps in each trajectory, the position sampling rate is at 10 Hz. The Argoverse dataset includes High Definition (HD) maps featuring the road centerlines and a rasterized drivable area map.

The dataset is divided into around 200,000 training samples, 40,000 validation samples, and 80,000 testing samples. In each sequence, the first 20 samples are observations and the following 30 need to be predicted. In each scene, a single agent is handpicked to be challenging and only its prediction is evaluated. We chose this to be the anchor agent, and locate the scene coordinate system at the last observed step.

Multimodal predictions are evaluated using four metrics:
\begin{itemize}
    \item \textbf{min Final Displacement Error (minFDE)} is the minimum L2 distance between the last predicted position of all the modes and the GT position.
    \item \textbf{min Average Displacement Error (minADE)} is the average L2 distance between the GT trajectory and the predicted mode with the minimum FDE.
    \item \textbf{Miss Rate (MR)} is the rate between the number of samples for which the minFDE is less than 2~m and the total number of GT trajectories. 
    \item \textbf{Drivable Area Compliance (DAC)} is the rate between the number of trajectories that remain inside the road and the total number of trajectories.
\end{itemize}

\subsection{Results}
\label{section:NA}

We measure the performance of our architecture on the Argoverse test set and compare it with state-of-the-art methods in Table \ref{tab:SOTA}. We show that each recent method outperforms by a wide margin the Nearest Neighbor (NN) technique proposed as an Argoverse baseline \cite{argoverse}. Our method is able to outperform the winners of the Argoverse Forecasting Challenge 2019 (SAMMP \cite{jean2019multihead} and UULM-MRM) on both minFDE and MR. Furthermore, our method performs on par with the other published state-of-the-art methods and with the winner of the Argoverse Forecasting Challenge 2020. This demonstrates the effectiveness of the proposed modules and of \emph{explicit training}, and that applying contextual information at the decoder stage is an effective solution.

\subsection{Ablation Study}
\label{section:TM}

\begin{table}[th!]
\begin{center}
\begin{tabular}{c p{0.15cm}p{0.15cm}p{0.15cm}p{0.15cm}p{0.15cm}p{0.3cm} cccc}
Fig
&\mcrot{1}{l}{55}{Social Interaction}
&\mcrot{1}{l}{55}{Map Information}
&\mcrot{1}{l}{55}{Decoder Map Addition}
&\mcrot{1}{l}{55}{Soft WTA}
&\mcrot{1}{l}{55}{Random Transformations}
&\mcrot{1}{l}{55}{Explicit Training}
&\mcrot{1}{l}{55}{\textbf{minADE}}
&\mcrot{1}{l}{55}{\textbf{minFDE}}
&\mcrot{1}{l}{55}{\textbf{MR}}
&\mcrot{1}{l}{55}{\textbf{DAC}}\\
\hline
\ref{fig:s5}
&\Checkmark
&\Checkmark
&\Checkmark
&\Checkmark
&\Checkmark
&\Checkmark
& 0.809 & 1.108 & 0.101 & 0.988\\
\ref{fig:s6}
&\Checkmark
&\Checkmark
&\Checkmark
&\Checkmark
&\Checkmark
&
& 0.921 & 1.192 & 0.119 & 0.984\\
\ref{fig:s7}
&\Checkmark
&\Checkmark
&\Checkmark
&\Checkmark
&
&
& 1.001 & 1.401 & 0.167 & 0.984\\
\ref{fig:s8}
&\Checkmark
&\Checkmark
&\Checkmark
&
&
&
& 1.138 & 1.756 & 0.230 & 0.978\\
\hline
\ref{fig:s1}
&\Checkmark
&\Checkmark
&\Checkmark
&\Checkmark
&\Checkmark
&
& 0.921 & 1.192 & 0.119 & 0.984\\
\ref{fig:s2}
&\Checkmark
&\Checkmark
&
&\Checkmark
&\Checkmark
&
& 0.940 & 1.286 & 0.147 & 0.973\\
\ref{fig:s3}
&\Checkmark
&
&
&\Checkmark
&\Checkmark
&
& 0.983 & 1.501 & 0.184 & 0.974\\
\ref{fig:s4}
&
&
&
&\Checkmark
&\Checkmark
&
& 1.053 & 1.787 & 0.232 & 0.955\\
\end{tabular}
\end{center}
 \caption{Ablation study on the modules used in the network and the training techniques.}
 \label{tab:ablation}
\end{table}

After analyzing the performance achieved with respect to the state-of-the-art, in Table \ref{tab:ablation} we show the effect of each module and of the training techniques studying performance on the Argoverse validation set. We evaluate the effect of training techniques by progressively ablating them from the training of our proposed network. Table \ref{tab:ablation} shows that all training techniques are beneficial for the performance of the network. They are particularly valuable as they allow to improve the performance of the model without requiring additional complexity at forecasting time.

When studying the effect of the modules used in the network, we apply soft WTA and random rotations but not \emph{explicit training}. This is done to fairly evaluate the contribution of each module independently of the training technique, as the model including social interaction and map information benefits more from \emph{explicit training}. Compared to the proposed network, we first move the agent-to-lane attention to the encoder to show the benefit of a decoder-centered architecture and then remove the map inclusion and the social interaction layer. We can see that adding social interaction on top of the basis model can significantly improve performance. The use of map information in the encoder can further improve the behavior but a stronger boost is achieved when the map information is fused with social interaction inside the decoder. This supports our claim that adding information about the road structure can be particularly effective if done inside the decoder.

In Figure \ref{fig:ablation}, we show the qualitative performance of each method evaluated in the ablation study on a challenging forecasting scenario. This is a hard example as the observed trajectory is the one of an agent standing still in front of a junction. Therefore, the prediction needs to be formulated only based on the contextual information and the interaction with the other visible agent. Our proposed decoder-oriented network - Figure \ref{fig:s5} - shows the best understanding of the driving scenario as it generates two sets of trajectories accounting for the two possible driving decisions. It is also visible how the progressive ablation of training techniques and network modules deteriorates the ability to predict realistic trajectories, down to the basis network in Figure \ref{fig:s4}, which forecasts driving in a straight line.

\section{CONCLUSIONS}

In our work, we present a novel approach to motion forecasting focused on exploiting the interaction with the environment as the trajectory is predicted. We propose a network using task specific modules and exploiting map information only at the decoder stage and a set of techniques, like \emph{explicit training} and soft WTA, designed to properly train it. We show through experiments that our method is able to perform on par with the current state-of-the-art and we analyse the effect of each contribution. In future work we plan to integrate our decoder-focused networks with a planning module, where the decoder-focused architecture should allow for a tighter integration. 

\addtolength{\textheight}{-0.1cm}   

\bibliographystyle{plain}
\bibliography{./root.bbl}

\end{document}